# Robust Facial Landmark Detection under Significant Head Poses and Occlusion

Yue Wu     Qiang Ji

ECSE Department, Rensselaer Polytechnic Institute

110 8th street, Troy, NY, USA

{wuy9,jiq}@rpi.edu

## Abstract

*There have been tremendous improvements for facial landmark detection on general "in-the-wild" images. However, it is still challenging to detect the facial landmarks on images with severe occlusion and images with large head poses (e.g. profile face). In fact, the existing algorithms usually can only handle one of them. In this work, we propose a unified robust cascade regression framework that can handle both images with severe occlusion and images with large head poses. Specifically, the method iteratively predicts the landmark occlusions and the landmark locations. For occlusion estimation, instead of directly predicting the binary occlusion vectors, we introduce a supervised regression method that gradually updates the landmark visibility probabilities in each iteration to achieve robustness. In addition, we explicitly add occlusion pattern as a constraint to improve the performance of occlusion prediction. For landmark detection, we combine the landmark visibility probabilities, the local appearances, and the local shapes to iteratively update their positions. The experimental results show that the proposed method is significantly better than state-of-the-art works on images with severe occlusion and images with large head poses. It is also comparable to other methods on general "in-the-wild" images.*

## 1. Introduction

Facial landmark detection refers to the localization of the fiducial points on facial images. With the detected points, human facial shape and appearance information can be utilized for facial analysis. Recently, there are tremendous improvements of the facial landmark detection algorithms on general "in-the-wild" images (Figure 1(a)). However, it is still challenging to detect the facial landmarks on images with severe occlusion and large head poses (e.g. pose > 60 degree, self-occlusion)(Figure 1(b)(c)).

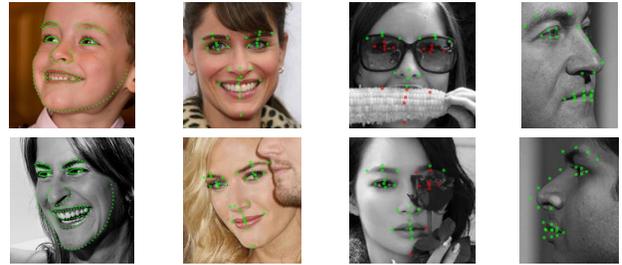

(a) General "in-the-wild" images    (b) occlusion    (c) Profile face

Figure 1. Predicted landmark locations and occlusions (red dots in (c) indicate occluded points) with the proposed method. Images are from Helen [16], LFPW [2], COFW [4], and FERET [19] sets.

The existing algorithms usually can only handle either images with occlusion [4][11][28][10][14] or images with large head poses [27][30]. In addition, they treat them differently. For example, pose dependent [30] or occlusion dependent [4][28] models are trained to handle different cases. However, if we regard the landmarks on the self-occluded facial parts as occluded points, where the face itself is the occluder, we can consider images with large head poses as special cases of images with occlusion and treat them similarly. Based on this intuition, we propose a novel method to handle both images with severe occlusion and images with large head poses.

The general framework of the proposed robust cascade regression method is shown in Figure 2 and 3. First, we initialize the landmark locations using the mean face shape and assume all the points are visible in the first iteration. Then, to achieve robustness, instead of directly predicting the binary landmark occlusion vectors and landmark locations, we gradually update the landmark visibility probabilities and locations iteratively in a coarse to fine manner. When updating the visibility probability, we utilize the appearance and shape information that depend on the currently estimated landmark locations. In addition, we explicitly add occlusion pattern as a constraint. When updating the landmark locations, we consider the appearance and shape information with weights that depend on the landmark visibility probabilities. The landmark locations and visibility probabilities interact to reach convergence. We highlight the major contributions of the proposed work:

- **General framework:** The proposed method is the first algorithm that can handle both images with severe occlusion and images with large head poses. It treats self-occlusion in images with large head poses as a special case of image occlusion.
- **Occlusion prediction:** Our occlusion prediction method is different from the previous works [4][28]. While they train several occlusion dependent models (e.g. mouth is occluded), we handle them in one unified framework. In addition, we explicitly add occlusion pattern as a constraint.
- **Landmark localization:** For facial landmark detection with occlusion, we treat points differently based on their visibility probabilities. We explicitly add the shape features for prediction. For images with large head poses, since the landmark annotations are missing on the self-occluded facial parts, we propose a learning method to handle this issue. The facial shape pattern is implicitly embedded in the model.
- **Experimental results:** The proposed method performs well on general "in-the-wild" images, and it is significantly better than the other state-of-the-art works on images with severe occlusion and large head poses.
- **Database:** We annotated some images with large head poses from FERET database [19]. [1]

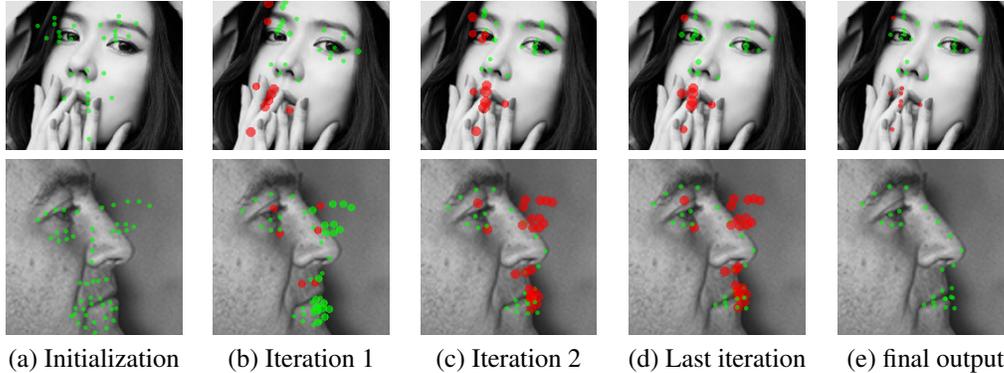

(a) Initialization (b) Iteration 1 (c) Iteration 2 (d) Last iteration (e) final output

Figure 2. Facial landmark detection and occlusion prediction in different iterations. First row: image with severe occlusion. Second row: image with large head pose. In (a)-(d), the radius of point is inversely proportional to the landmark visibility probabilities and the point is marked with red color if its visibility probability is smaller than a threshold. In (e), for images with large head pose, we only show the visible points as the final output. (Better see in color)

The remaining part of the paper is organized as follows. In section 2, we review the related work. In section 3, we introduce the proposed method. In section 4, we discuss the experiments and we conclude the paper in section 5.

[1] Landmark annotations can be downloaded: http://www.ecse.rpi.edu/~cvrl/wuy/FERET_annotation.rar

## 2. Related Work

Facial landmark detection algorithms can be classified into three major categories, including the holistic methods, the constrained local methods and the regression based methods. The proposed method follows the regression framework, but it is specifically designed to handle occlusion and large head poses.

The holistic methods build global appearance and shape models during learning and fit testing image by estimating the model parameters. The differences among the holistic methods lie in the fitting procedure and they usually follow either the least squares formulation [17][1] or the linear regression formulation [6].

The constrained local methods [8] combine global face shape model and local appearance model for facial landmark detection. This approach is usually superior to the holistic methods, since it relies on more flexible local appearance model that is easier to learn. Typical constrained local methods usually focus on the representations of the face shapes [2][24] or the local appearances [22][7].

Recently, the regression based methods show more promising performance than the holistic methods and the constrained local methods. Unlike the methods in the other two categories, the regression based methods do not explicitly build the global appearance or shape models. Instead, they directly map the local facial appearance to the landmark locations. For example, the absolute coordinates of the facial landmarks can be estimated directly from the facial appearance with the conditional regression forests [9] or deep convolutional neural networks [23]. Different from [9][23], most of the other regression based methods [26][20][5][15] start from an initial face shape, and they gradually update the landmark locations based on the local appearances. For those regression based methods, cascade techniques are usually embedded in the framework to improve both the robustness and accuracy. One limitation of the regression based methods is that they usually can only be trained with fully supervised data. Therefore, they cannot learn one unified model to handle both frontal face and profile face with missing landmark location annotations. In this work, we improve the method to solve this issue.

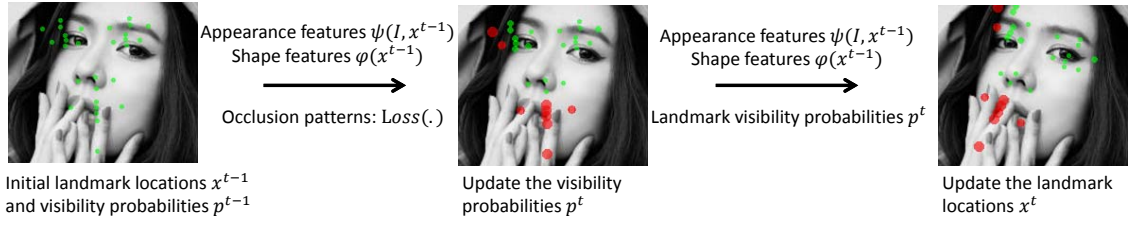

Figure 3. Facial landmark detection in one iteration. (Better see in color)

Despite the fact that facial landmark detection accuracy has been dramatically improved on general "in-the-wild" images, facial landmark detection remains challenging for facial images with severe occlusion or large head poses. There are only a few algorithms focusing on those challenging cases. For example, for images with severe occlusion, occlusion dependent models are built based on the partial appearances from pre-defined non-occluded facial parts for landmark detection and occlusion estimation in [4] and [28]. During testing, the detection results from all occlusion dependent models are merged together with weights depending on the landmark occlusion prediction results. In [11], a probabilistic graphical model is built to infer the landmark occlusions and locations jointly. For images with large head poses, there are even fewer works [30][27]. In [30], several models are built to handle facial landmark detection with different pre-defined discrete head poses. The detection result with the highest score is outputted as the final result. In [27], 3D model is used to handle images with large head poses. Different from the previous works, we propose a general framework to handle both cases and we do not build pose dependent or occlusion dependent models.

## 3. Approach

### 3.1. The general framework

The goal of the facial landmark detection algorithm is to find out the mapping from image $\mathbf{I}$ to landmark locations $\mathbf{x} \in \Re^{2D_l}$, where $D_l$ is the number of facial landmarks. To handle images with occlusion and large head poses, we introduce the landmark visibility probability variable $\mathbf{p} \in [0,1]^{D_l}$. For one specific image with binary landmark occlusion vector $\mathbf{c} \in \{0,1\}^{D_l}$, $p_d$ measures the probability that a landmark is visible ($c_d = 1$).

The general framework of the proposed robust cascade regression method is shown in Figure 2, 3 and Algorithm 1. With the initial landmark locations and the assumption that all the landmarks are visible at the beginning, the algorithm

**Algorithm 1**: The general framework

Initialize the landmark locations $\mathbf{x}^0$ using the mean face;
Assume all the landmarks are visible $\mathbf{p}^0 = \mathbf{1}$
**for** *t=1, 2, ..., T or convergence* **do**

    Update the landmark visibility probabilities given the images, the current landmark locations, and the occlusion pattern $Loss(.)$.
$$f_t : \mathbf{I}, \mathbf{x}^{t-1}, Loss(.) \to \Delta \mathbf{p}^t$$
$$\mathbf{p}^t = \mathbf{p}^{t-1} + \Delta \mathbf{p}^t$$

    Update the landmark locations given the images, the current landmark locations, and the landmark visibility probabilities.
$$g_t : \mathbf{I}, \mathbf{x}^{t-1}, \mathbf{p}^t \to \Delta \mathbf{x}^t$$
$$\mathbf{x}^t = \mathbf{x}^{t-1} + \Delta \mathbf{x}^t$$

**end**
Output the estimated landmark locations $\mathbf{x}^T$ and the binary occlusion vector based on the predicted visibility probabilities $\mathbf{p}^T$.

updates the visibility probabilities and the landmark locations across iterations to achieve convergence. When updating the visibility probabilities, we introduce a constrained supervised regression model, denoted as $f_t$, to predict the landmark visibility probability update $\Delta \mathbf{p}^t$ based on the image, the current landmark locations $\mathbf{x}^{t-1}$ and the occlusion pattern embedded in a loss function $Loss(.)$. When updating the landmark locations, we use a regression function $g_t$ that predicts the landmark location update $\Delta \mathbf{x}^t$ based on the image, the current landmark locations $\mathbf{x}^{t-1}$, and the visibility probabilities $\mathbf{p}^t$. In the following, we discuss each part in details.

### 3.2. Update the landmark visibility probability

The landmark visibility probability and landmark occlusion are difficult to predict. First, there are large variations of the facial appearance on the occluded facial parts, since the occlusion could be caused by arbitrary objects. Second, due to the poor landmark detection results in the first few iterations, direct occlusion prediction is infeasible. Thus, a better choice is to update the landmark visibility probabilities by accumulating information through iterations. Third, due to the occlusion consistency among nearby points, there exists certain occlusion pattern, which should be embedded in the model as a constraint. However, since any part of the face could be occluded by arbitrary objects, there are a large number of possible patterns and the occlusion patterns could be complex. Thus, it's not appropriate to pre-define the possible occlusion pattern (e.g. mouth is occluded) as the existing works [4][28], and more effective model should be used. Fourth, it's not enough to learn the occlusion pattern from limited training images with exhaustive human annotations. In fact, the occlusion pattern can be learned from synthetic data. Last but not the least, the regression function should depend on both the local appearances and the current shapes for joint prediction. Based on those intuitions, we propose to update the landmark visibility probabilities based on the appearance and shape information from all points and use the learned explicit occlusion pattern as a constraint.

### 3.2.1 Landmark visibility prediction model

Landmark visibility prediction depends on the local appearance, the current shape, and the occlusion pattern. To encode the appearance information, we use SIFT features of the local patches centered at the current landmark locations, denoted as $\phi(\mathbf{I}, \mathbf{x}^{t-1}) \in \Re^{D_l D_f}$ ($D_f$ =128 is the dimension of features). To encode the shape information, we calculate the differences of x, y coordinates for pairwise landmarks to get the shape features denoted as $\varphi(\mathbf{x}^{t-1}) \in \Re^{D_l(D_l-1)}$, which provide the scale, pose, and non-ridge information of the current face. By combining the appearance and shape information, we can generate a concatenated feature vector denoted as $\Psi(\mathbf{I}, \mathbf{x}^{t-1}) = [\phi(\mathbf{I}, \mathbf{x}^{t-1}); \varphi(\mathbf{x}^{t-1})]$. To encode the occlusion pattern, we learn a loss function $Loss(\mathbf{c})$ for occlusion vector $\mathbf{c}$ (there are $2^{D_l}$). The loss function penalizes the infrequent and infeasible occlusion label configurations (e.g. every other point is occluded). Then, we can update the landmark visibility probabilities $\mathbf{p}^t$ for the next iteration:

$$\begin{aligned}
\underset{\Delta \mathbf{p}^t}{\text{minimize}} \quad & \|\Delta \mathbf{p}^t - T^t \Psi(\mathbf{I}, \mathbf{x}^{t-1})\|_2^2 + \lambda \mathbb{E}_{\mathbf{p}^t}[Loss(\mathbf{c})] \\
\text{subject to} \quad & \mathbf{p}^t = \mathbf{p}^{t-1} + \Delta \mathbf{p}^t \\
& \mathbf{0} \leq \mathbf{p}^t \leq \mathbf{1}
\end{aligned} \quad (1)$$

$$\mathbb{E}_{\mathbf{p}^t}[Loss(\mathbf{c})] = \sum_{k=1}^{2^{D_l}} Loss(\mathbf{c}_k) P(\mathbf{c}_k; \mathbf{p}^t) \quad (2)$$

$$P(\mathbf{c}; \mathbf{p}) = \prod_{d=1}^{D_l} p_d^{c_d}(1-p_d)^{1-c_d} \quad (3)$$

In the first term of the objective function, we use linear regression function with parameter $T^t$ to predict the landmark visibility probability update $\Delta \mathbf{p}^t$ from the appearance and shape features $\Psi(\mathbf{I}, \mathbf{x}^{t-1})$, and we want to minimize the standard least squares error. In the second term, we want to minimize the expected loss of the occlusion vector, where the expectation is taken over the visibility probabilities $\mathbf{p}^t$ we want to infer for the next iteration. The basic idea is to minimize the discrepancy between the predicted binary occlusion vector and the prior occlusion pattern encoded in $Loss(\mathbf{c})$. The expectation is denoted as $\mathbb{E}_{\mathbf{p}^t}[Loss(\mathbf{c})]$ and it is detailed in Equation 2 and 3. $\lambda$ is the hyper-parameters. In the following, we first explain model learning, and then discuss model inference.

### 3.2.2 Learning the landmark visibility prediction model

Model learning refers to the estimation of the linear regression parameter $T^t$ for each iteration and the loss function $Loss(.)$, which should be learned before the cascade training.

We use the Autoencoder model [3] (Figure 4 (a)) to learn the loss function $Loss(.)$ that captures the prior occlusion pattern based on the landmark occlusion labels of the training data and the synthetic data. We generate the synthetic landmark occlusion labels by sampling different numbers of occluders (up to 4 rectangles with random sizes and locations) in the face region and superimposing them onto the mean face (see Figure 4 (b) for one example). Then, based on all the feasible landmark occlusion label $\mathbf{c}_i$ from the real training data and the synthetic data, we learn the Autoencoder model with parameters $\theta = \{\mathbf{W}_1, \mathbf{b}_1, \mathbf{W}_2, \mathbf{b}_2\}$ that can minimize the reconstruction errors:

$$\theta^* = arg \min \sum_i \|\mathbf{c}_i - \sigma(\mathbf{W}_2 \sigma(\mathbf{W}_1 \mathbf{c}_i + \mathbf{b}_1) + \mathbf{b}_2)\|_2^2, \quad (4)$$

where $\sigma(.)$ is the sigmoid function. The model is pre-trained with Restricted Boltzmann Machine model and fine-tuned jointly [13]. After model learning, the loss function is defined as the reconstruction errors $Loss(\mathbf{c}; \theta) = \|\mathbf{c} - \sigma(\mathbf{W}_2 \sigma(\mathbf{W}_1 \mathbf{c} + \mathbf{b}_1) + \mathbf{b}_2)\|_2^2$. Figure 4 (c) shows the distributions of reconstruction errors of the feasible occlusion labels for Auto-encoder training, and random binary data, consisting of both feasible and infeasible occlusion vectors. The reconstruction error apparently can penalize the random infeasible occlusion vectors.

For the estimation of linear regression function with parameter $T^t$ in each iteration, we use standard least squares formulation. Specifically, given the training image, the currently estimated landmark locations $\mathbf{x}_i^{t-1}$, we can calculate the appearance and shape features $\Psi(\mathbf{I}_i, \mathbf{x}^{t-1})$. By subtracting the currently estimated landmark visibility probabilities $\mathbf{p}_i^{t-1}$ from the ground truth probabilities $\mathbf{p}_i^*$, we can get the landmark visibility probability update $\Delta \mathbf{p}_i^t$. Then, $T^t$ could be estimated by the standard least-squares formulation with closed form solution.

$$T^{t*} = arg \min_{T^t} \sum_i \|\Delta \mathbf{p}_i^t - T^t \Psi(\mathbf{I}_i, \mathbf{x}_i^{t-1})\|_2^2 \quad (5)$$

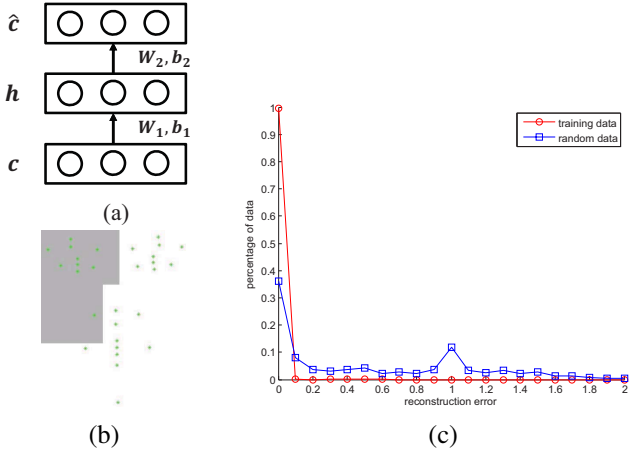

Figure 4. (a) Learning the occlusion patterns with Autoencoder. (b) Generate the synthetic occlusion labels. Green points: mean face shape. Gray areas: randomly generated occluders. (c) The distributions of reconstruction errors. Training data: feasible occlusion labels for Auto-encoder training. Random data: randomly generated binary data consisting of both feasible and infeasible occlusion vectors (e.g. every other point is occluded).

### 3.2.3 Inference with the landmark visibility prediction model

In inference, with Equation 1, we need to estimate $\Delta \mathbf{p}^t$ given the appearance and shape features $\Psi(\mathbf{I}, \mathbf{x}^{t-1})$, the currently estimated visibility probabilities $\mathbf{p}^{t-1}$, model parameter $T^t$, and the loss function $Loss(.)$. The inference is non-trivial, since the calculation of the expectation (Equation 2) would sum over $2^{D_l}$ possible occlusion vectors which would be intractable if the number of landmarks $D^l$ is large. To tackle this problem, we use Monte Carlo approximation and calculate the second term over the samples (K=5000 in our experiments).

$$\mathbb{E}_{\mathbf{p}^t}[Loss(\mathbf{c})] \approx \frac{const}{K} \sum_{k=1}^{K} Loss(\widetilde{\mathbf{c}}_k) P^t(\widetilde{\mathbf{c}}_k), \quad (6)$$

where $\widetilde{\mathbf{c}}_k$ is sampled by assuming all vectors are equally likely, and $const = 2^{D_l}$.

To solve the optimization problem in Equation 1 with the approximation in Equation 6, we use gradient descent algorithm. If we denote the gradient of the objective function w.r.t. $\Delta \mathbf{p}^t$ as $\delta$, then:

$$\delta = 2(\Delta \mathbf{p}^t - T^t \Psi(\mathbf{I}, \mathbf{x}^{t-1})) + \lambda \frac{const}{K} \sum_{k=1}^{K} Loss(\widetilde{\mathbf{c}}_k) \frac{\partial P^t(\widetilde{\mathbf{c}}_k)}{\partial \Delta \mathbf{p}^t}, \quad (7)$$

Given the gradient $\delta$, $\Delta \mathbf{p}^t$ can be updated by moving along the descent direction, with the constraint that $\mathbf{p}^t = \mathbf{p}^{t-1} + \Delta \mathbf{p}^t$ is in the range $[0, 1]^{D_l}$.

### 3.3. Update the landmark locations

Given the predicted landmark visibility probabilities, we need to update the landmark locations. There are a few issues and difficulties for landmark localization on images with occlusion and large head poses. First, the points should not be treated equally. For points with low visibility probabilities, the local appearances would be less useful and reliable, since the appearance of the occluder has limited information about the landmark locations. Second, for the detection of the occluded landmarks, the shape features and the shape constraint from other landmarks are more important than its local appearance. Third, due to the self-occlusion issue on images with large head poses, the location annotations for landmarks on the occluded facial parts are not available.

Based on the intuitions illustrated above, we modify the regression based method for landmark detection. Specifically, we predict the location update vector $\Delta \mathbf{x}^t$ with linear regression function as bellow:

$$\Delta \mathbf{x}^t = R^t [\sqrt{\mathbf{p}^t} \circ \Psi(\mathbf{I}, \mathbf{x}^{t-1})] \quad (8)$$

Here, "$\circ$" denotes the block-wise product between the square of the landmark visibility probabilities and the appearance features from corresponding point (we keep the shape features unchanged). In this case, the prediction will rely more on appearance features from points with high visibility probabilities, while treat the shape features equally. Since all points are estimated together starting from the mean face, the shape constraint is automatically embedded in learning. Thus, the algorithm may detect the occluded points based on the shape constraint. During detection, we could only output the landmark locations with high visibility probabilities ($> 0.6$) (second row of Figure 2 (e)).

In model training, we need to estimate the parameter $R^t$ in each iteration with missing landmark location annotations (e.g. no landmark annotations on self-occluded facial parts). To tackle this incomplete annotation issue, for each training image, we introduce the binary variable $\mathbf{w} \in \{0, 1\}^{D_l}$ to indicate whether the location annotation of a specific landmark is available ($w_k = 1$) or not ($w_k = 0$). Then, combining $\mathbf{w}_i$, the location update $\Delta \mathbf{x}_i^t$ (subtracting the current landmark locations $\mathbf{x}_i^{t-1}$ from the ground truth $\mathbf{x}_i^*$), the currently estimated visibility probability $\mathbf{p}_i^t$, and the concatenated appearance and shape features $\Psi(\mathbf{I}_i, \mathbf{x}_i^{t-1})$, parameter learning can be formulated as a weighted least squares problem with closed form solution.

$$R^t = arg\min_{R^t} \sum_i \|\Delta \mathbf{x}_i^t - R^t[\sqrt{\mathbf{p}_i^t} \circ \Psi(\mathbf{I}_i, \mathbf{x}_i^{t-1})]\|_{diag(\mathbf{w}_i)}^2, \quad (9)$$

where $diag(\mathbf{w}_i) \in \Re^{2D_l \times 2D_l}$ is a diagonal matrix and the corresponding elements (for x,y coordinates) are 0 if the landmark location annotation is missing. Therefore, parameter learning for the corresponding rows of $R^t$ will not depend on the specific data with missing landmark annotation.

### 3.4. Discussion

**Differences with Supervised Descent method (SDM) [26]:** 1) SDM learns the descent direction for facial landmark detection. It is not designed to handle occlusions. 2) SDM cannot handle images with large head pose with severe self-occlusion.[2] 3) The derivation in [26] shows that the regression function should change according to the current shape, while SDM fix it as constant. In the proposed method, our shape features compensate this limitation.

**Differences with Robust Cascade Pose Regression method (RCPR)** [4]: 1) RCPR builds several occlusion dependent models which only draw features from 1/9 of the facial region, assuming the region is not occluded. Based on the limited information from 1/9 of the facial region, it's difficult for RCPR to predict the landmark locations and their occlusions on the other 8/9 facial region. In addition, the pre-defined 9 models can not effectively cover all possible occlusion cases. On the contrary, the proposed method trains one unified model, which is more flexible and robust. It draws features from the whole regions and considers them with different weights, which is more effective. 2) RCPR cannot handle images with large head poses. 3) There is no explicit occlusion pattern in RCPR.

**Differences with Face detection, Pose estimation, Landmark Localization algorithm (FPLL)** [30]: 1) FPLL follows the Constrained Local Method. It builds several pose dependent models and chooses the detection result with the highest score from all models. It would lead to poor result if the model is chose incorrectly. On the contrary, we propose a unified model that automatically solve the problem. 2) FPLL can not handle images with severe occlusion.

## 4. Experimental results

In this section, we evaluate the proposed method on images with severe occlusion, images with large head poses, and general "in-the-wild" images.

### 4.1. Implementation details

**Databases:** We use three kinds of databases. The first kind of databases contain general "in-the-wild" images collected from the internet with near-frontal head poses (less than 60 degree) and limited occlusion (about 2%). They are the Labeled Face Parts in the Wild (LFPW) database [2] with 29 points and the Helen database [16] with 194 points. For LFPW database, due to the invalid URLs, we only collected 608 images for training and 152 images for testing from the internet. For Helen database, following the previous works [16][20], we use 2000 images for training and use the remaining 330 images for testing. The second kind of database contains "in-the-wild" images with severe occlusion (about 25%). Here, we use the Caltech Occluded Faces in the Wild (COFW) database [4]. There are 1345 images for training and 507 images for testing. The annotations include the landmark locations and the binary occlusion labels for 29 points. The third kind of database contains images with large head poses (e.g. profile face, head pose larger than 60 degrees). Most of the facial images comes from the MultiPIE database [12] and the Facial Recognition Technology (FERET) database [19]. The annotations of 39 visible points on MultiPIE are provided by the database and [30]. We annotated 11 points on profile images from FERET. Sample images from different databases can be found in Figure 8.

**Evaluation criteria:** Following the previous works, we calculate the error as the distance between detected landmarks and the ground truth landmarks normalized by the inter-ocular distance. For the third kind of database with profile images, we normalized the error by half of the distance between the outer eye corner and mouth corner. Throughout the paper, we calculate the average error across all available annotated landmarks from the testing databases.

**Model parameters:** When calculating the SIFT features, the radius of the local image patch is about 0.14 of the face size. There are 4 cascade iterations for the model. To augment the training images, following the previous works, we perturb the scale, rotation angle, and position of the initial face shape for parameter learning. The hyper-parameter $\lambda$ in Equation 1 is 0.001. We use Autoencoder with one hidden layer and we set 20 and 25 hidden nodes for experiments in section 4.2 and 4.3, respectively.

### 4.2. Images with severe occlusion

In this section, we show the performance of the algorithms on the challenging COFW database with severe occlusion (about 25%). For fair comparison, following the previous work [4], the algorithm is trained with training set from COFW and tested on the testing set. For the proposed method, we implemented three versions and they are denoted as ours_baseline (no shape features as discussed in section 3.2.1, no occlusion pattern constraint as defined in the second term of Equation 1), ours_ShapeFea (with shape features, no occlusion pattern), and ours_Full (with shape features and occlusion pattern).

**Facial landmark detection:** The experimental results are shown in Table 1 and Figure 8 (a). The proposed method is close to human performance and it is significantly better than all the other state-of-the-art works, including the Cascaded Regression Copse (CRC) [10], the Occlusion Coherence (OC) [11], SDM [26], RCPR [4], Explicit Shape Regression (ESR) [5], and FPLL algorithm [30]. Note that for the algorithms that can perform well on general "in-the-wild" databases, such as SDM [26] and ESR [5], there are significant performance drops on the COFW database with severe occlusion. In addition, the proposed method is significantly better than SDM that ignores the occlusion pattern and shape features. Comparing three versions of the proposed method, we see that the shape features, and occlusion pattern are important for good performance.

---
[2]Global SDM [25] solves this issue to some extent.

Table 1. Comparison of facial landmark detection errors and occlusion prediction results on COFW database (29 points) [4]. The reported results from the original papers are marked with "*".

| algorithm | Landmark detection error | | Occlusion prediction |
|---|---|---|---|
| | visible points | all points | precision/recall% |
| Human | - | 5.6 [4] | - |
| CRC [10] | - | 7.30* | - |
| OC [11] | - | 7.46* | 80.8/37.0%* |
| SDM [26] | 6.69 | 7.70 | - |
| RCPR [4] | - | 8.50* | 80/40%* |
| ESR [5] | - | 11.20 | - |
| FPLL [30] | - | 14.40 | - |
| ours_baseline | 5.68 | 6.54 | 80/43.78% |
| ours_ShapeFea | 5.22 | 5.96 | 80/46.07% |
| ours_Full | 5.18 | 5.93 | 80/49.11% |

**Occlusion prediction:** The occlusion prediction results are shown in Table 1 and Figure 8 (a). Following the previous work [4], we fix the precision to be about 80%, and compare the recall values. As can be seen, the proposed method is much better than OC [11] and RCPR [4], which are the state-of-the-art works.

**Performance across iterations:** In Figure 5, we show the landmark detection errors and the recall values (fixing precision as 0.8, following [4]) based on the estimated occlusion probabilities across all four iterations using the proposed method (ours_Full). As can be seen, both the landmark detection and the occlusion prediction results improve over iterations and they converge quickly.

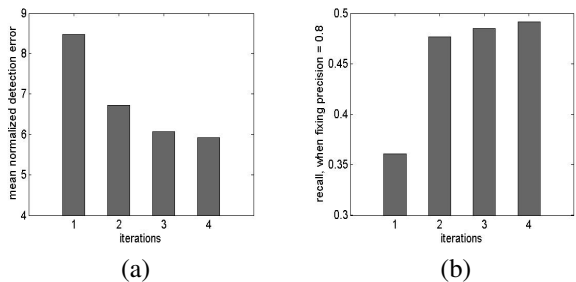

(a) (b)

Figure 5. Performance of the proposed method across iterations on COFW database. (a) Landmark detection errors. (b) Occlusion prediction accuracy (recall values at precision=0.8).

### 4.3. Images with large head poses

In this section, we evaluate the full model of the proposed method and compare it to other algorithms on challenging images with large head poses (larger than 60 degrees). To the best of our knowledge, the FPLL algorithm [30] and Pose-free algorithm [27] are the only two methods that can perform facial landmark detection on images with large head poses, due to the self-occlusion issue. However, exact fair comparison to them is infeasible, since they are trained on the combinations of different subsets of MultiPie databases [12] and other databases. To ensure relatively fair comparison, we use the software provided by the authors for FPLL and pose-free algorithm, and we train the proposed method with similar procedure. For training, we use the MultiPie database with 14 poses, the training set from Helen and LFPW databases with 51 landmarks. We test all the algorithms on profile faces from FERET database (244 right profile face, 221 left profile face).

The experimental results are shown in Figure 6, 7 and 8 (b). In Figure 6, we plot the cumulative error distribution curves. In Figure 7, for different algorithms, we show the images with the largest fitting errors. As can be seen, the proposed method is significantly better than the FPLL algorithm [30] and Pose-free algorithm [27] on the profile faces. The Pose-free algorithm could result in large facial landmark detection error if the estimated pose is wrong (e.g. Figure 7(a)). Among all the 465 testing images, our algorithm only fail to predict the correct landmark occlusions on one image (last image in Figure 8(b)).

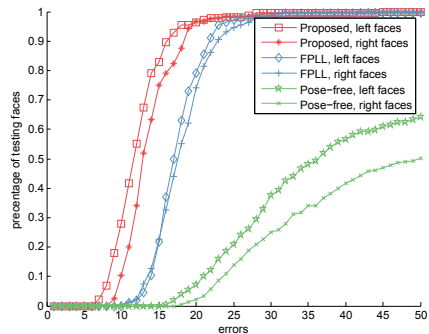

Figure 6. Cumulative error distribution curves on profile faces.

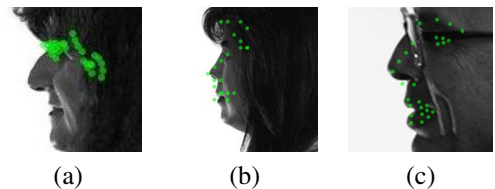

(a) (b) (c)

Figure 7. For different methods, we show the images with the worst fitting results. (a) Pose-free [27], (b) FPLL [30], (c) proposed method.

### 4.4. General "in-the-wild" images

Finally, we evaluate the proposed method on general and less challenging "in-the-wild" images and compare it to more state-of-the-art works. Note that, most of the algorithms that perform well on general "in-the-wild" images do not work well on challenging images with severe

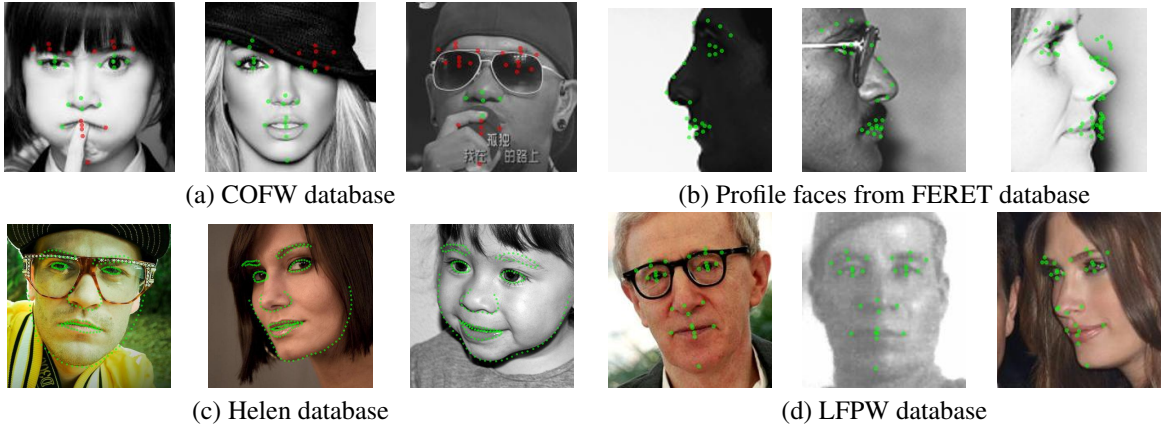

(a) COFW database  (b) Profile faces from FERET database

(c) Helen database  (d) LFPW database

Figure 8. Facial landmark detection results with the proposed method on four databases. (a) COFW [4] database with severe occlusion. Red points: the proposed method predicts them as occluded landmarks. (b) Profile faces from FERET database [19]. We show all points that the algorithm predicted as visible points. (c)(d) General "in-the-wild" Helen [16] and LFPW [2] database. (Better see in color)

occlusion or large head poses. Following the previous works [16][20][26], for each database, we use the training set to learn the model and test it on the testing set.

The experimental results on Helen and LFPW databases are shown in Table 2 and Figure 8 (c)(d). The results on Helen database show that the performance of the proposed method is better than the other state-of-the-art works, including the fast version of Local Binary Feature (LBF) method [20], SDM [26], RCPR [4], ESR [5], the CompASM [16], and the Extended Active Shape Model (STASM) [18]. For the experiments on LFPW database, the training and testing data varies from method to method. We only can get access to half of the training data comparing to the data used in the original Consensus of exemplars (COE) [2]. However, the performance of the proposed method is still comparable to the other methods. We also tested the proposed method on the most challenging ibug set [21] with 135 images. Our method achieves the detection error of 11.52, which is better than the state-of-the-art works [20][26][5], among which the best algorithm achieves error of 11.98.

The speed of the proposed method is comparable to other state-of-the-art works. For the model without the explicit occlusion pattern, the proposed algorithm can achieves 15 frames per second running on a single core machine with matlab implementation. With the occlusion pattern, the full model of the proposed algorithm achieves 2 frames per second. This is comparable to some state-of-the-art methods (e.g. [26][4]), but slower than the others (e.g. [20]). But, again, those fast algorithms may not work well on images with severe occlusion or large head poses.

## 5. Conclusion

In this work, we propose a general facial landmark detection algorithm to handle images with severe occlusion and images with large head poses. We iteratively update the landmark visibility probabilities and landmark locations. For occlusion prediction, we train one unified model to handel different kinds of occlusion and we explicitly add the prior occlusion pattern as the constraint. For landmark detection, we treat points differently and rely more on the information from points with high visibility probabilities. The experimental results show that the proposed method is significantly better than the other state-of-the-art works on images with severe occlusion and images with large head poses. It is also comparable to other methods on general and less challenging "in-the-wild" images.

Table 2. Comparison of facial landmark detection errors on Helen database [16] (194 points) and LFPW database (29 points) [2]. The reported results from the original papers are marked with "*".

| algorithm | Helen | LFPW |
|---|---|---|
| LBF [20] | **5.41*** (fast 5.80*) | **3.35***(fast 3.35*) |
| SDM [26] | 5.82 | 3.47* |
| RCPR [4] | 6.50* | 3.50* |
| ESR [5] | 5.70* | 3.5* |
| EGM [29] | - | 3.98* |
| COE [2] | - | 3.99* |
| OC [11] | - | 5.07* |
| CompASM [16] | 9.10* | - |
| STASM [18] | 11.1 | - |
| FPLL [30] | - | 10.91 |
| ours | **5.49** | 3.93 |
|  |  | (less data) |

In the future, we would further improve the algorithm in two directions. First, we would extend the detection algorithm for realtime tracking. Second, we would improve the algorithm so that it can handle more challenging cases in real world conditions (e.g. significant illumination change, low resolution, etc.).